\documentclass[runningheads]{llncs}
\usepackage[T1]{fontenc}
\usepackage{graphicx}

\usepackage{multirow}
\usepackage{tabularray}
\usepackage{booktabs}
\usepackage[hidelinks]{hyperref}
\usepackage{amsmath}
\usepackage{makecell}
\usepackage[symbol]{footmisc}

\newsavebox\CBox
\newcommand*\textBF[1]{\sbox\CBox{#1}\resizebox{\wd\CBox}{\ht\CBox}{\textbf{#1}}}

\usepackage{color}

\begin{document}
\title{Facial Wrinkle Segmentation for Cosmetic Dermatology: Pretraining with Texture Map-Based Weak Supervision}
\titlerunning{Facial Wrinkle Segmentation}
%
\author{Junho Moon\inst{1}\orcidID{0009-0004-3522-6357} \and
Haejun Chung$^*$\inst{1}\orcidID{0000-0001-8959-237X} \and
Ikbeom Jang\thanks{Corresponding authors}\inst{2}\orcidID{0000-0002-6901-983X}}
\authorrunning{J. Moon et al.}
%
\institute{Hanyang University, Seoul 04763, Republic of Korea \\
\email{\{jhmoon6807, haejun\}@hanyang.ac.kr} \and
Hankuk University of Foreign Studies, Yongin 17035, Republic of Korea\\
\email{ijang@hufs.ac.kr}}
\maketitle              
\begin{abstract}
Facial wrinkle detection plays a crucial role in cosmetic dermatology. Precise manual segmentation of facial wrinkles is challenging and time-consuming, with inherent subjectivity leading to inconsistent results among graders. To address this issue, we propose two solutions. First, we build and release the first public facial wrinkle dataset, `FFHQ-Wrinkle', an extension of the NVIDIA FFHQ dataset. It includes 1,000 images with human labels and 50,000 images with automatically generated weak labels. This dataset could serve as a foundation for the research community to develop advanced wrinkle detection algorithms. Second, we introduce a simple training strategy utilizing texture maps, applicable to various segmentation models, to detect wrinkles across the face. Our two-stage training strategy first pretrain models on a large dataset with weak labels (N=50k), or masked texture maps generated through computer vision techniques, without human intervention. We then finetune the models using human-labeled data (N=1k), which consists of manually labeled wrinkle masks. The network takes as input a combination of RGB and masked texture map of the image, comprising four channels, in finetuning. We effectively combine labels from multiple annotators to minimize subjectivity in manual labeling. Our strategies demonstrate improved segmentation performance in facial wrinkle segmentation both quantitatively and visually compared to existing pretraining methods. The dataset is available at \url{https://github.com/labhai/ffhq-wrinkle-dataset}.

\keywords{Facial wrinkle segmentation \and Weakly supervised learning \and Texture map pretraining \and Transfer learning}
\end{abstract}
\section{Introduction}
\begin{figure}[tb]
  \centering
  \includegraphics[width=1.0\textwidth]{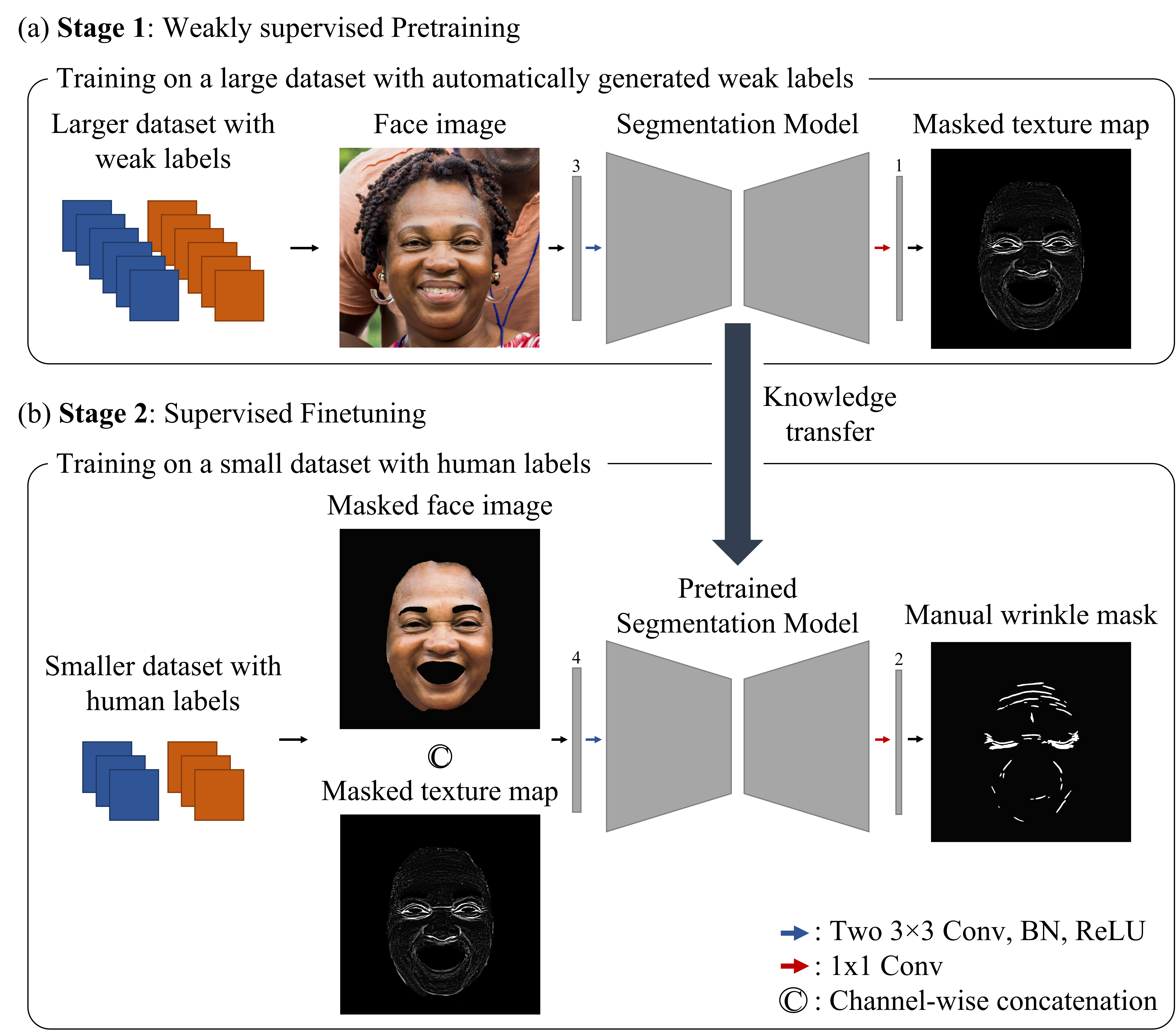}
  \caption{Two-stage training for facial wrinkle segmentation. (a) Weakly supervised pretraining stage: the model learns to extract masked texture maps from RGB face images. (b) Supervised finetuning stage: the model refines its ability to extract facial wrinkles from RGB-masked face images and masked texture maps. The model parameters are initialized with the weights from the weakly supervised pretraining stage.}
  \label{fig:fig6}
\end{figure}

With the growing interest in dermatological diseases and skin aesthetics, predicting facial wrinkles is becoming increasingly significant. Facial wrinkles serve as critical indicators of aging~\cite{old2010aznar2010much,old2010luu2010combined,old2018ng2018hybrid}, and are essential for evaluating skin conditions~\cite{skinCondition1991warren1991age,skinDiagnosis2009kim2009wrinkle}, diagnosing dermatological disorders~\cite{skinDiagnosis2015wilder2015stimulated}, and planning pre-treatment protocols for skin management~\cite{skinPreTreatment2008allemann2008hyaluronic,skinPreTreatment2019satriyasa2019botulinum}. Nevertheless, the manual detection of facial wrinkles poses considerable challenges. Accurate detection and analysis of facial wrinkles necessitate a high level of expertise, typically available only through well-trained professionals such as dermatologists. This process is time-consuming and entails substantial costs due to the extensive time and effort required by the experts.

Recently, numerous studies have focused on the automatic segmentation of facial wrinkles through the application of deep learning techniques~\cite{sabina2021edge,sabina2021nasolabial,kim2022semi,kim2023facial,chen2023facial,yang2024striped}. Nevertheless, these deep learning-based approaches are notably data-intensive. Due to the intricate distribution of facial wrinkles across the face, analyzing extensive collections of images can be exceedingly resource-intensive if each wrinkle must be individually evaluated. Furthermore, the manual analysis procedure is fraught with subjectivity. The assessments of individual experts can differ significantly based on their experience, level of training, and personal biases, thereby complicating the consistency and reproducibility of the analysis results.

To address these challenges, we propose a two-stage training strategy, as illustrated in Fig.~\ref{fig:fig6}. This approach utilizes computer vision techniques, specifically filters, to generate many weakly labeled wrinkle masks (N=50,000) without human intervention for weakly supervised pretraining. A smaller set of accurately labeled wrinkle masks (N=1,000) is employed for supervised finetuning. This method significantly decreases the time and cost associated with manual wrinkle labeling, providing substantial advantages over traditional methodologies.
To ensure the development of a generalized and robust model, we conducted experiments using a dataset comprising images captured from various angles, lighting conditions, races, ages, and skin conditions. We quantitatively analyzed the challenges associated with consistent manual wrinkle labeling across such a diverse dataset and integrated data labeled by multiple annotators to reduce subjectivity during the finetuning stage. No public dataset exists for full-face wrinkle segmentation, although there are a few private datasets. To address this gap, we have made our dataset publicly accessible to enhance the reproducibility and reliability of our results. This initiative aims to reduce the manual labeling costs for future research and serve as a benchmark dataset.

\section{Related works}

\subsection{Deep learning-based facial wrinkle segmentation}
Deep learning-based methods for facial wrinkle segmentation aim to enable neural network models to learn the features necessary for accurate wrinkle detection autonomously. Kim et al.~\cite{kim2022semi} introduced a semi-automatic labeling strategy to enhance performance by extracting texture maps from face images and combining them with roughly labeled wrinkle masks, utilizing a U-Net architecture~\cite{unet2015ronneberger2015u} for segmentation. In a subsequent study~\cite{kim2023facial}, they further improved segmentation accuracy by implementing a weighted deep supervision technique, which employs a weighted wrinkle map to more precisely calculate the loss for the downsampled decoder, outperforming traditional deep supervision methods. Yang et al.~\cite{yang2024striped} developed Striped WriNet, which integrates a Striped Attention Module composed of Multi-Scale Striped Attention and Global Striped Attention within a U-shaped network. This approach applies an attention mechanism across multiple scales, effectively segmenting both coarse and fine wrinkles.

\subsection{Weakly supervised learning}
Weakly supervised learning is a methodology that trains models using incomplete or inaccurate labeled data instead of fully labeled data in situations where strong supervision information is lacking~\cite{wsl2018zhou2018brief}. Xu et al.~\cite{wsl2019xu2019camel} proposed CAMEL, a weakly supervised learning framework that uses a MIL-based label expansion technique to divide images into grid-shaped instances and automatically generate instance-level labels, enabling histopathology image segmentation with only image-level labels. Shen et al.~\cite{wsl2019ji2019scribble} trained a deep learning model using only scribbles on whole tumors and healthy brain tissue, along with global labels for the presence of each substructure, to segment all sub-regions of brain tumors.

\section{Dataset}

\begin{figure}[tb]
  \centering
  \includegraphics[width=0.95\textwidth]{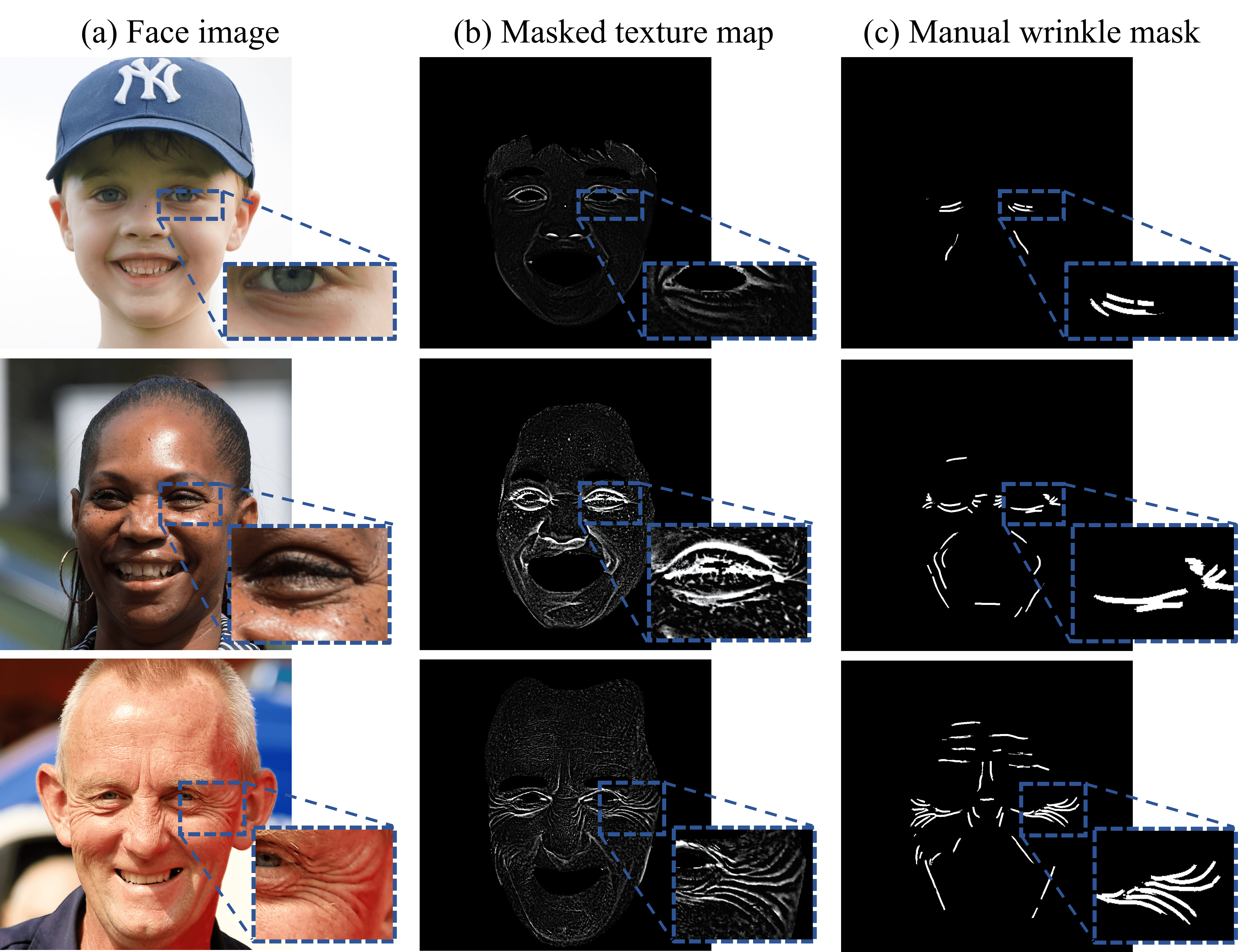}
  \caption{Training Dataset. (a) High-resolution face images. (b) Masked texture maps extracted from face images, which include information about facial features. (c) Reliable manual wrinkle masks created by combining the results of multiple annotators.}
  \label{fig:fig1}
\end{figure}

\subsection{Dataset specifications}
The first public facial wrinkle dataset, `FFHQ-Wrinkle', comprises pairs of face images and their corresponding wrinkle masks. We focused on wrinkle labels while utilizing the existing face image dataset FFHQ (Flickr-Faces-HQ)~\cite{ffhq2019karras2019style}, which contains 70,000 high-resolution (1024x1024) face images captured under various angles and lighting conditions. The dataset we provide consists of one set of manually labeled wrinkle masks (N=1,000) and one set of `weak' wrinkle masks, or masked texture maps, generated without human labor (N=50,000). We selected 50,000 images from the FFHQ dataset, specifically image IDs 00000 to 49999. We used these 50,000 face images to create the weakly labeled wrinkles and randomly sampled 1,000 images from these to create the ground truth wrinkles. The methods for generating weakly labeled wrinkles and ground truth wrinkles are discussed in Section \ref{Training strategy}. Table~\ref{tab:dataset_config} summarizes estimated demographic information of the dataset--i.e. age, race, and sex. The age and sex data were sourced from the FFHQ-Aging~\cite{metadataor2020lifespan} dataset, where at least three annotators labeled each image. The race/ethnicity attribute was obtained through facial attribute analysis using the DeepFace\footnote{\url{https://github.com/serengil/deepface}} framework. Hence, the demographic information may include errors. As illustrated in Fig.~\ref{fig:fig1}, the dataset consists of individuals of varying ages, sex, and race/ethnicity, featuring a range of skin conditions such as freckles, acne, and pigmentation. This diversity makes the dataset particularly suitable for training models to handle the wide array of skin conditions encountered in clinical settings. The dataset is publicly available at \url{https://github.com/labhai/ffhq-wrinkle-dataset}.

\begin{table}[tb]
\centering
\caption{Demographic attributes of the dataset. The `Human-labeled' data represents the 1,000 face images manually labeled by human annotators and the `Weakly-labeled' data refers to the 50,000 images labeled without human intervention.}
\label{tab:dataset_config}
\resizebox{\textwidth}{!}{
\begin{tabular}{c|c|cc}
\toprule
\multicolumn{2}{c|}{Dataset} & Human-labeled & Weakly-labeled \\ \hline
\multicolumn{2}{c|}{Sample size} & 1000 & 50000 \\ \hline
\multirow{2}{*}{Age} & 0-9 / 10-19 / 20-29 / & 66 / 68 / 233 / & \, 7030 / 4448 / 13804 / \\
    & 30-39 / 40-49 / 50-69 / 70+ & \, 246 / 186 / 161 / 40 \, & \, 10960 / 6931 / 5550 / 1277 \, \\ \hline
Sex & Male / Female & 471 / 529 & 26929 / 23071 \\ \hline
\multirow{2}{*}{\, \makecell{Race/ \\ Ethnicity} \,} &\, White / Asian / Latino Hispanic / \,& 587 / 210 / 67 / & 29728 / 11121 / 3895 / \\ 
            & Black / Middle Eastern / Indian & 81 / 37 / 18 /  & 2383 / 2053 / 820 \\
\bottomrule
\end{tabular}}
\end{table}

\subsection{Ground truth wrinkle annotation} \label{Ground truth wrinkle generation}
\begin{figure}[t]
  \centering
  \includegraphics[width=0.9\textwidth]{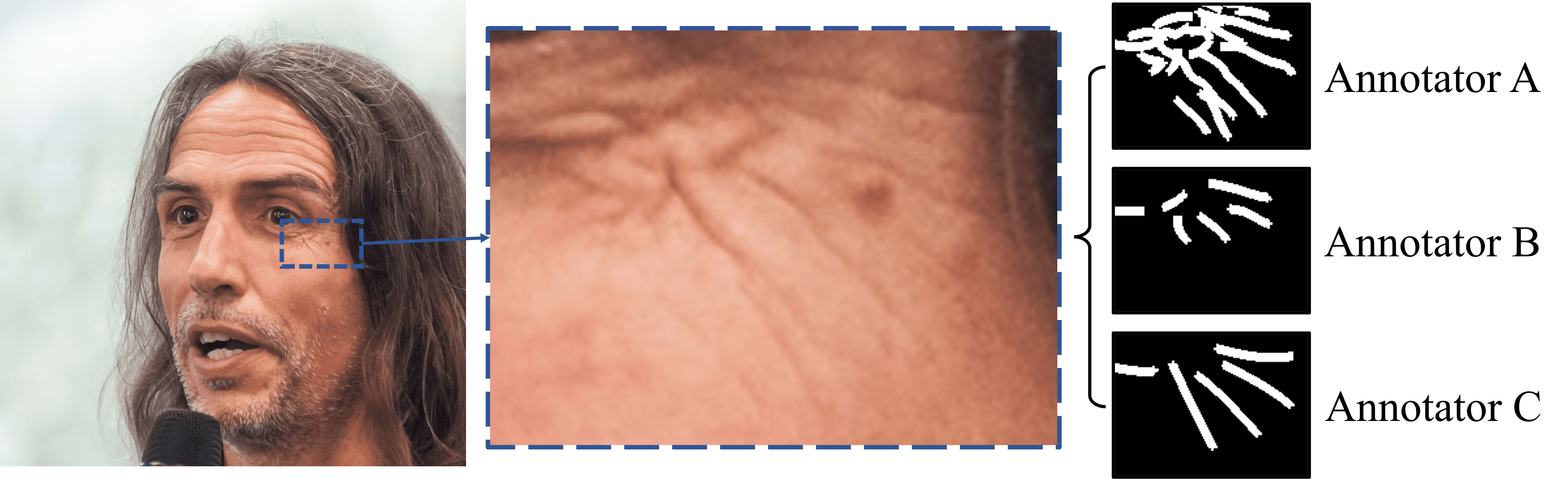}
  \caption{Ambiguity in wrinkle evaluation. The labeling results from three annotators for the same image are different.}
  \label{fig:fig4}
\end{figure}

For ground truth wrinkles, we manually annotated the face images. The annotation process involved three annotators with extensive experience in image processing and analysis. Wrinkles can be categorized into two types—dynamic wrinkles and static wrinkles~\cite{wrinkletypewu1996simulation}. Dynamic wrinkles are formed by facial muscles and appear with expressions but disappear when the face is at rest. Static (permanent) wrinkles are visible even when the face is at rest and result from the repeated formation of dynamic wrinkles over time. We annotated both types of wrinkles without distinguishing between them. Given the subjectivity inherent in wrinkle data, a consistent standard for wrinkle assessment was established prior to the commencement of labeling. The annotators conducted three synchronization sessions to minimize inter-rater variability. The annotation primarily targeted the forehead, crow's feet, and nasolabial folds, encompassing the overall facial area. Due to the high resolution and diversity of the dataset—comprising various races, skin conditions, backgrounds, and angles—achieving consistent labeling results proved challenging, even with established standards for wrinkle assessment, as illustrated in Fig. \ref{fig:fig4}. Consequently, as demonstrated in Table \ref{tab:tab1}, the inter-rater agreement was low, underscoring the highly subjective nature of wrinkle assessments.

\begin{table}[t]
\centering
\caption{Inter-rater agreement of manual wrinkle annotation. The Jaccard similarity index and Pearson correlation coefficient between different annotators are analyzed.}
\label{tab:tab1}
\resizebox{\textwidth}{!}{ 
\begin{tabular}{c|c c c c}
\toprule
Metric & \, Annotators A\&B \, & \, Annotators B\&C \, & \, Annotators A\&C \, & Average \, \\ \hline  
Jaccard similarity index & 0.2631 & 0.2962 & 0.3182 & 0.2925 \, \\  
\, Pearson correlation coefficient \,& 0.4167 & 0.4559 & 0.4928 & 0.4551 \, \\
\bottomrule
\end{tabular}}
\end{table}

\section{Method}
\subsection{Model architecture}
We evaluated our proposed method using the U-Net~\cite{unet2015ronneberger2015u} and Swin UNETR~\cite{swinunetr} architectures, with U-Net serving as the base model for ablation studies and additional experiments. As depicted in Fig. \ref{fig:fig6}, the U-Net model features a standard architecture comprising four encoder blocks and four decoder blocks. The Swin UNETR model employs an encoder with a window size of 16 and patches of size 4x4, projecting the input patch into a 48-dimensional embedding space. This model includes four encoder blocks, each consisting of two successive Swin Transformer blocks~\cite{swintliu2021swin}, and four decoder blocks.

\subsection{Training strategy}
\label{Training strategy}
We train the segmentation model using a substantial number of masked texture maps in a weakly supervised manner, followed by finetuning with a smaller set of reliably manually labeled wrinkle masks in a supervised manner. This training strategy, which involves finetuning the weights of a pretrained model that extracts facial textures using human-labeled wrinkle data, significantly enhances the model's capability to detect facial wrinkles. The overall training pipeline is illustrated in Fig. \ref{fig:fig6}.

\subsubsection{Weakly supervised pretraining stage}
\begin{figure}[tb]
  \centering
  \includegraphics[width=0.9\textwidth]{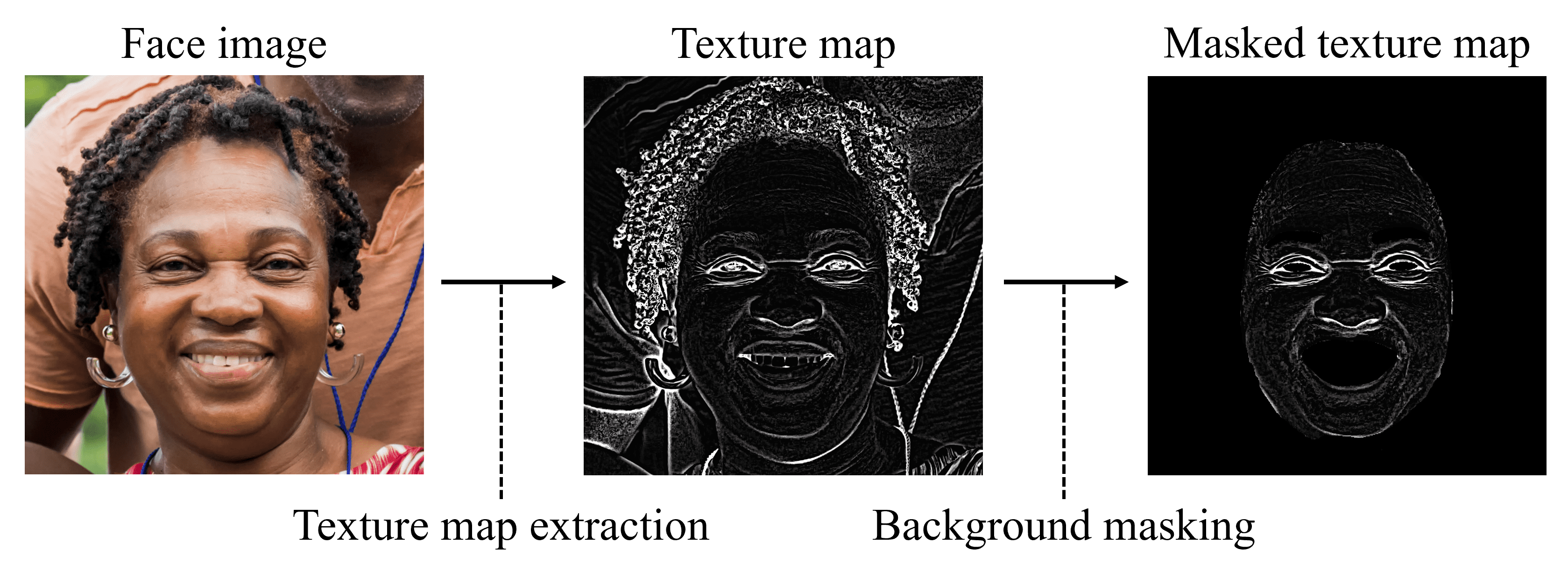}
  \caption{Weakly labeled wrinkle generation pipeline. After extracting the texture map from the face image, we mask the non-facial regions to generate a masked texture map containing information on facial features. This masked texture map is then used as a weakly labeled wrinkle.}
  \label{fig:weak_wrinkle}
\end{figure}

In the pretraining stage, we utilized weakly labeled wrinkle data automatically extracted through computer vision techniques without human intervention as the ground truth. Fig. \ref{fig:weak_wrinkle} illustrates the pipeline for generating weakly labeled wrinkles for the weakly supervised pretraining stage. Utilizing Equation (1), we extracted the texture map~\cite{kim2022semi} from the face image through a Gaussian kernel-based filter.

\begin{equation}
     T(x,y)=(1-\frac{I(x,y)}{1+I_{G(\sigma)}(x,y)})\times 255
\end{equation}
where $G$ represents the Gaussian kernel, $\sigma$ denotes its standard deviation, $I_{G(\sigma)}$ is the Gaussian filtered image, and $(x,y)$ are the pixel coordinates in the image. Following the methodology in~\cite{kim2022semi}, we set the Gaussian kernel's standard deviation to 5 and its size to 21x21 for texture map extraction. The extracted texture map contains detailed information about the contours, curves, and skin textures of the face image. However, as the texture map includes numerous false positives from the background, we employ a BiSeNet~\cite{bisenet2018yu2018bisenet} architecture-based facial parsing deep learning model\footnote{\url{https://github.com/zllrunning/face-parsing.PyTorch}} to mask non-facial regions, resulting in the final masked texture map used as ground truth. We avoid converting the masked texture map into a binary mask due to the variability in the size, shape, and depth of wrinkles, which makes determining an appropriate threshold challenging. Fig. \ref{fig:fig1}-(b) shows the masked texture map used as the final ground truth in the weakly supervised pretraining stage.

In the weakly supervised pretraining stage, the model takes a 3-channel RGB face image as input and outputs a 1-channel masked texture map (Fig. \ref{fig:fig6}-(a)). We use mean squared error (MSE) loss~\cite{mseloss1994nix1994estimating} to optimize the model, calculated as shown in equation (2). 
\begin{equation}
        MSE(\hat{y}, y)=\frac{1}{n}\sum_{i=1}^{n}\left(\hat{y_i}-y_i\right)^{2}
\end{equation}
where $\hat{y_i}$ and $y_i$  are the model output and the masked texture map, respectively.

\subsubsection{Supervised finetuning stage}
\begin{figure}[tb]
  \centering
  \includegraphics[width=0.95\textwidth]{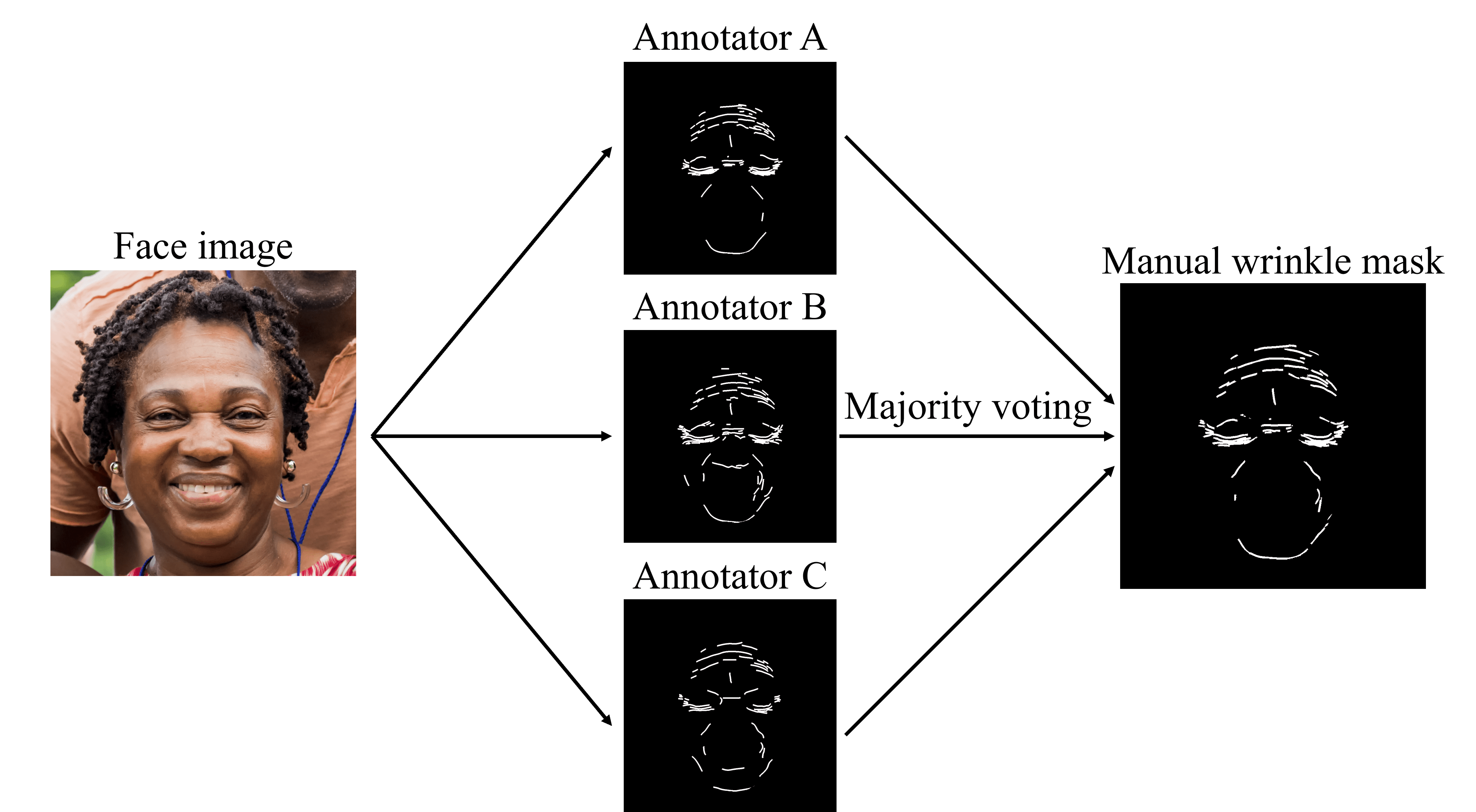}
  \caption{Ground truth wrinkle generation pipeline. We combine data labeled by multiple annotators through majority voting to create a reliable ground truth wrinkle.}
  \label{fig:fig2}
\end{figure}

For the ground truth in the finetuning stage, we utilized human-labeled wrinkle data generated as described in Section \ref{Ground truth wrinkle generation}. Fig.~\ref{fig:fig2} illustrates the pipeline of the ground truth generation of the wrinkle mask. To produce a reliable ground truth wrinkle mask, we used majority voting to retain only the pixels that were labeled by at least two groups, thereby reducing variability among the annotators. Fig. \ref{fig:fig1}-(c) displays the manual wrinkle mask used as the final ground truth in the supervised finetuning stage. As model inputs, we use masked face images, where non-facial regions were masked using a facial-parsing model. Additionally, we included masked texture maps, which were used as ground truth in the pretraining stage, as auxiliary inputs.

In the supervised finetuning stage, the model takes as input a 3-channel RGB face image with only the facial regions and a 1-channel masked texture map. It then produces a 2-channel output indicating the presence of wrinkles and background. This stage begins with the model parameters from the pretraining stage, where the model was weakly supervised to extract masked texture maps from face images. Using transfer learning, we refine the model by adjusting its weights with manually labeled wrinkle masks. This process enhances the model's ability to detect facial wrinkles by building on the general facial texture extraction skills developed during pretraining. We optimize the model using soft Dice loss~\cite{diceloss2006crum2006generalized}, as shown in equation (3).
\begin{equation}
    DL(p, g) = 1 - \frac{1}{C}\sum_{c=1}^{C}\frac{2\sum_{i=1}^{N}p_{i,c}g_{i,c}}{\sum_{i=1}^{N}p_{i,c}+\sum_{i=1}^{N}g_{i,c}}
\end{equation}
where $C$ is the total number of classes, $N$ is the total number of pixels, $p_{i,c}$ represents the predicted probability for pixel $i$ belonging to class $c$, and ${g_{i,c}}$ represents the ground truth label for pixel $i$ belonging to class $c$, respectively.

\section{Experiments}

\subsection{Implementation details} \label{Implementation details}
In both the weakly supervised pretraining and supervised finetuning stages, we utilize the original 1024x1024 image-label pairs as inputs without resizing. The AdamW optimizer~\cite{adamw2017loshchilov2017decoupled} is employed, configured with a weight decay of 0.05, $\beta_{1}$ set to 0.9, and $\beta_{2}$ set to 0.999. We also implement the SGDR scheduler~\cite{sgdr2016loshchilov2016sgdr}. To maintain dataset diversity, we randomly apply various augmentations, including horizontal flipping, scaling, affine transformation, elastic transformation, grid distortion, and optical distortion during training. The dataset is partitioned into 80\% for training, 10\% for validation, and 10\% for testing.

\subsubsection{Weakly supervised pretraining stage}
In the weakly supervised pretraining stage, the model is trained for 300 epochs. The SGDR scheduler starts with an initial period of 100 epochs, with the learning rate beginning at a maximum of 0.001 and decaying to 0 over the period. At the end of each period, the length of the next period doubles that of the previous one. The batch size is 26 for U-Net and 22 for Swin UNETR. All pretraining processes were performed on an NVIDIA A100 Tensor Core GPU.

\subsubsection{Supervised finetuning stage}
In the supervised finetuning stage, the U-Net model is finetuned for 150 epochs, while the Swin UNETR model is finetuned for 300 epochs. The batch size is 14 for both models. The SGDR scheduler's initial period length is set to 50 epochs for U-Net and 100 epochs for Swin UNETR. The learning rate starts at a maximum of 0.0001 and decreases to 0 within each period. At the end of each period, the length of the next period doubles that of the previous one, with the maximum learning rate set to 90\% of the last period's maximum. All finetuning processes are performed on RTX A6000 and RTX 6000 Ada GPUs.

\subsection{Evaluation metrics}
To evaluate the performance of the final finetuned model in wrinkle segmentation, we use the Jaccard Similarity Index (JSI), F1-score, and Accuracy (Acc). 

The Jaccard Similarity Index measures the overlap between the predicted wrinkle regions and the ground truth regions, defined as follows:
\begin{equation}
     \text{JSI}=\frac{|A \cap B|}{|A \cup B|}
\end{equation}
where $A$ is the predicted segmentation, and $B$ is the actual label.

The F1-score is the harmonic mean of precision and recall, while accuracy measures the proportion of correctly predicted pixels out of the total pixels. They are defined as follows:
\begin{equation}
    \text{Precision} = \frac{TP}{TP + FP}
\end{equation}
\begin{equation}
    \text{Recall} = \frac{TP}{TP + FN}
\end{equation}
\begin{equation}
    \text{F1-score} = 2 \times \frac{\text{Precision} \times \text{Recall}}{\text{Precision} + \text{Recall}}
\end{equation}
\begin{equation}
    \text{Acc} = \frac{TP + TN}{TP + TN + FP + FN}
\end{equation}
where $TP$ is the number of true positives, $FP$ is the number of false positives, $FN$ is the number of false negatives, and $TN$ is the number of true negatives.

\subsection{Results}
To evaluate the performance of our proposed method, we first compare it with the latest methods: the semi-automatic labeling and weighted deep supervision method~\cite{kim2023facial}, and the Striped WriNet method~\cite{yang2024striped}. Because the primary contribution of this work is the pretraining strategy, we also compare it with other pretraining techniques. They include using ImageNet pretrained models and self-supervised learning methods. For the ImageNet pretrained models, we replace the encoder part of the U-shape architecture with models pretrained on the ImageNet-1K dataset~\cite{imagenetrussakovsky2015imagenet}; specifically, we use ResNet-50~\cite{residualhe2016deep} for U-Net and Swin-T~\cite{swintliu2021swin} for Swin UNETR. For the self-supervised learning methods, we use denoising self-supervised learning~\cite{denoising2022brempong2022denoising} for pretraining U-Net, setting the Gaussian distribution’s standard deviation to 0.2, and masked image prediction~\cite{mim2022xie2022simmim} for pretraining Swin UNETR, using 32x32 masked patches and a 60\% masking ratio. All training hyperparameters follow those specified in Section \ref{Implementation details}. To assess performance in scenarios with very limited labeled data, we train our model on the full training set (100\%, N=800) and on a randomly sampled subset (5\%, N=40).

\begin{figure}[tb]
  \centering
  \includegraphics[width=1.0\textwidth]{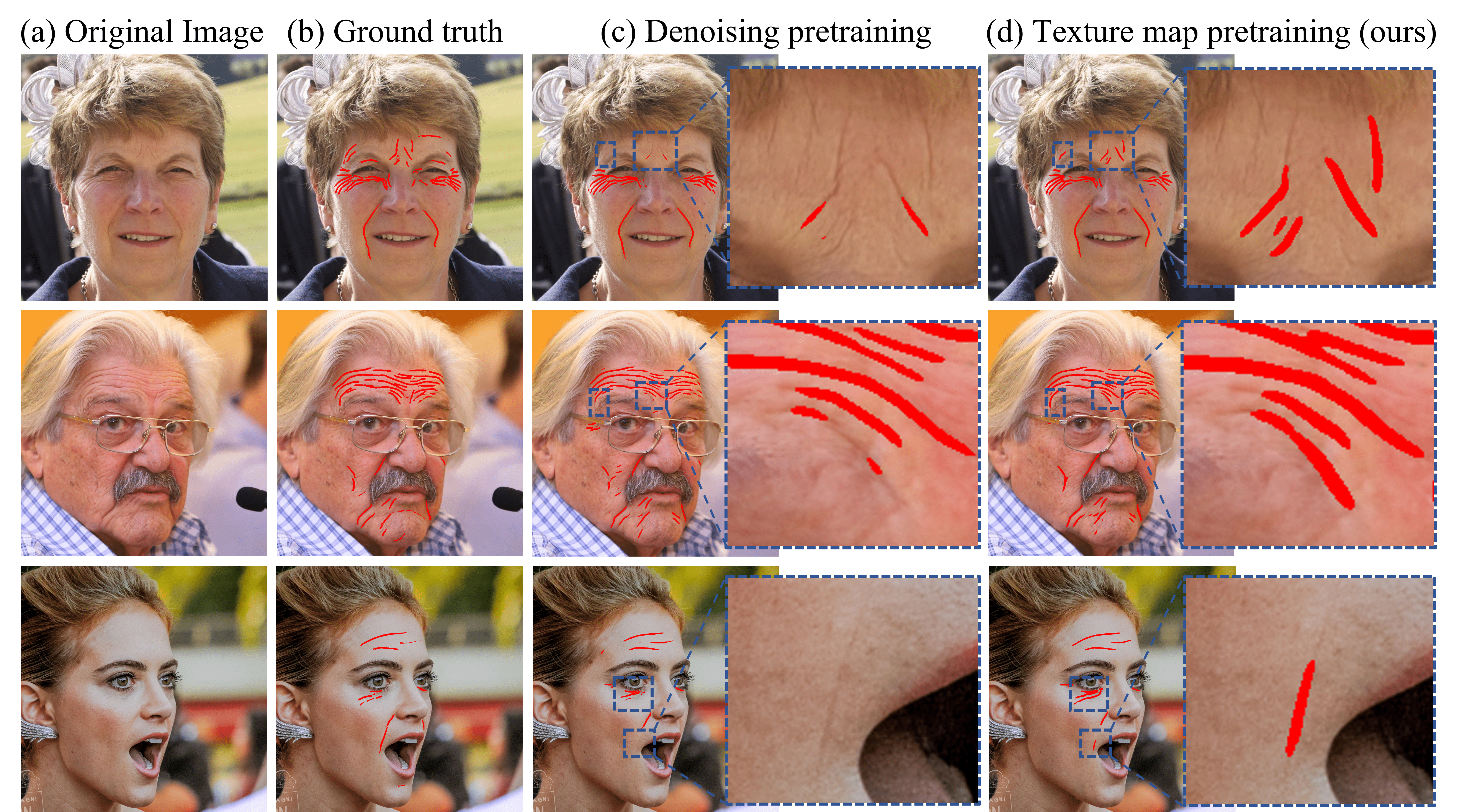}
  \caption{Qualitative comparison against the denoising pretraining method. The blue boxes highlight areas with significant visual differences. (a) Face image. (b) Ground truth wrinkle. (c) Predicted wrinkles from a model using self-supervised learning with denoising pretraining, followed by finetuning with a manual wrinkle mask. (d) Predicted wrinkles from our model, trained with weak supervision using a masked texture map and then finetuned with a manual wrinkle mask.}
  \label{fig:fig8}
\end{figure}

The proposed method outperforms the latest wrinkle segmentation methods and the ones using the same model architectures with different pertaining methods. The performance gap is much larger in data-limited situations—i.e., fine-tuned on 5\% of the manually-labeled data. Table \ref{tab:tab2} shows quantitative comparisons of wrinkle segmentation performance for each method using U-Net and Swin UNETR architectures. Our method consistently achieves the highest performance across both datasets and architectures. Fig. \ref{fig:fig8} presents a qualitative comparison of our method with denoising pretraining using U-Net, which is the next best performing method in experiments using 100\% of the data.

\subsection{Ablation study}
Incorporating the masked texture map as an additional input during the finetuning stage led to significant improvements in wrinkle segmentation, demonstrating the effectiveness of our approach. Table \ref{tab:tab3} presents quantitative comparisons using the U-Net architecture to assess the benefits of including a 1-channel masked texture map as an additional input during finetuning. We compare our pretraining method (Texture map pretraining) with a conventional approach (No pretraining), which is trained solely on manually labeled data, both with (RGB+Texture) and without (RGB) the additional masked texture map input.

\begin{table}[tb]
\centering
\caption{Quantitative comparisons of facial wrinkle segmentation performance. Our method is compared against two latest wrinkle segmentation methods, models trained without pretraining, and models using different pretraining strategies. These pretraining techniques include masked image prediction, denoising, and pretraining encoders using the ImageNet-1K dataset.}
\label{tab:tab2}
\resizebox{\textwidth}{!}{
\begin{tabular}{cc|ccc|ccc|c}
\toprule
\multicolumn{2}{c|}{\multirow{2}{*}{Method}} & \multicolumn{3}{c|}{100\% (N=800)} & \multicolumn{3}{c|}{5\% (N=40)} & \multirow{2}{*}{$n_{\text{params}}$} \\ \cline{3-8} 
\multicolumn{2}{c|}{} & \multicolumn{1}{c}{JSI} & \multicolumn{1}{c}{F1-score} & Acc & \multicolumn{1}{c}{JSI} & \multicolumn{1}{c}{F1-score} & Acc & \\ \hline
\multicolumn{2}{c|}{Semi automatic labeling + WDS~\cite{kim2023facial}} & \multicolumn{1}{c}{0.4552} & \multicolumn{1}{c}{0.6256} & 0.9954 & \multicolumn{1}{c}{0.3384} & \multicolumn{1}{c}{0.5057} & 0.9928 & 17.269M \\ \hline
\multicolumn{2}{c|}{Striped WriNet~\cite{yang2024striped}} & \multicolumn{1}{c}{0.4665} & \multicolumn{1}{c}{0.6294} & 0.9956 & \multicolumn{1}{c}{0.2382} & \multicolumn{1}{c}{0.3761} & 0.9903 & 6.223M \\ \hline
\multicolumn{1}{c|}{\multirow{4}{*}{\makecell{Swin UNETR \\with pretraining}}} & No pretraining & \multicolumn{1}{c}{0.4220} & \multicolumn{1}{c}{0.5858} & 0.9949 & \multicolumn{1}{c}{0.2545} & \multicolumn{1}{c}{0.3944} & 0.9932 & 25.153M \\ 
\multicolumn{1}{c|}{} & \makecell{ImageNet-1K~\cite{imagenetrussakovsky2015imagenet} \\ (Swin-T~\cite{swintliu2021swin})} & 0.4385 & 0.6028 & 0.9952 & {0.2877} & {0.4351} & 0.9939 & 100.56M \\  
\multicolumn{1}{c|}{} & Masked image modeling~\cite{mim2022xie2022simmim} & \multicolumn{1}{c}{0.4450} & \multicolumn{1}{c}{0.6079} & 0.9954 & \multicolumn{1}{c}{0.2963} & \multicolumn{1}{c}{0.4452} & 0.9937 & 25.153M \\ 
\multicolumn{1}{c|}{} & \textBF{Texture map (ours)} & \multicolumn{1}{c}{0.4643} & \multicolumn{1}{c}{0.6271} & 0.9953 & \multicolumn{1}{c}{0.3416} & \multicolumn{1}{c}{0.4970} & \textBF{0.9944} & 25.155M \\ \hline
\multicolumn{1}{c|}{\multirow{4}{*}{\makecell{U-Net \\ with pretraining}}} & No pretraining & \multicolumn{1}{c}{0.4638} & \multicolumn{1}{c}{0.6278} & 0.9955 & \multicolumn{1}{c}{0.3021} & \multicolumn{1}{c}{0.4551} & 0.9918 & 17.263M \\  
\multicolumn{1}{c|}{} & \makecell{ImageNet-1K~\cite{imagenetrussakovsky2015imagenet} \\ (ResNet-50~\cite{residualhe2016deep})} & {0.4664} & {0.6296} & 0.9955 & 0.3428 & 0.5018 & 0.9934 & 32.521M \\  
\multicolumn{1}{c|}{} & Denoising~\cite{denoising2022brempong2022denoising}& \multicolumn{1}{c}{0.4709} & \multicolumn{1}{c}{0.6339} & 0.9955 & \multicolumn{1}{c}{0.2840} & \multicolumn{1}{c}{0.4338} & 0.9898 & 17.263M \\ 
\multicolumn{1}{c|}{} & \textBF{Texture map (ours)} & \multicolumn{1}{c}{\textBF{0.4831}}& \multicolumn{1}{c}{\textBF{0.6442}} & \textBF{0.9957} & \multicolumn{1}{c}{\textBF{0.3512}} & \multicolumn{1}{c}{\textBF{0.5116}} & 0.9929 & 17.264M \\ 
\bottomrule
\end{tabular}}
\end{table}

\begin{table}[tb]
\centering
\caption{Ablation study of the effectiveness of adding a masked texture map as an additional model input. We conduct experiments using U-Net. The segmentation performance improves when using the masked texture map as an additional input during finetuning after texture map training.}
\label{tab:tab3}
\resizebox{\textwidth}{!}{
\begin{tabular}{c|c|ccc|ccc}
\toprule
\multirow{2}{*}{Method}         & \multirow{2}{*}{Model input} & \multicolumn{3}{c|}{100\% (N=800)} & \multicolumn{3}{c}{5\% (N=40)} \\ \cline{3-8} 
                                &                              & JSI      & F1-score     & Acc     & JSI     & F1-score    & Acc    \\ \hline
\multirow{2}{*}{No pretraining} & RGB (3-ch) & 0.4638 & 0.6278 & 0.9955 & 0.3021 & 0.4551 & 0.9918 \\ 
& RGB+Texture (4-ch) & 0.4606 & 0.6221 & 0.9954 & 0.3208 & 0.4743 & 0.9924 \\ \hline
\multirow{2}{*}{\textBF{\makecell{Texture map \\ pretraining}}} & RGB (3-ch) & 0.4796 & 0.6422 & \textBF{0.9957} & 0.3442 & 0.5051 & 0.9919 \\
& \textBF{RGB+Texture (4-ch, ours)} & \textBF{0.4831} & \textBF{0.6442} & \textBF{0.9957} & \textBF{0.3512} & \textBF{0.5116} & \textBF{0.9929} \\
\bottomrule
\end{tabular}}
\end{table}

\section{Discussion}
Our approach achieves state-of-the-art performance when compared to two publicly released models specifically designed for wrinkle segmentation, in addition to outperforming ImageNet pretrained models and self-supervised learning methods. We demonstrate that our two-stage training strategy significantly enhances wrinkle segmentation efficiency. Furthermore, our approach shows the potential to achieve high performance with limited data, which could enhance scalability and flexibility in clinical settings. By using a large amount of weakly labeled data obtained automatically through filters for weakly supervised training and then finetuning with a small amount of reliable manually labeled data, we significantly reduce the time and cost required for manual labeling while improving the segmentation performance of facial wrinkles. To minimize subjectivity in the manual labeling process, we effectively combine data labeled by multiple annotators, resulting in more reliable training data. Additionally, to enhance the reproducibility of our research and reduce the manual labeling costs for subsequent studies, we release the dataset publicly available, which can also serve as a benchmark dataset for future research. The performance improvement of facial wrinkle segmentation through transfer learning has not been conducted in previous research, indicating that our approach can be efficiently integrated into various tasks related to facial wrinkle detection and segmentation tasks. Additionally, since this research falls under the broader category of thin object detection tasks, it is expected to be widely applicable to studies requiring segmentation of thin objects (e.g., fundus imaging, vascular imaging).

According to our experimental results, the performance of the Swin UNETR, a hybrid transformer-CNN architecture, is lower compared to the standard CNN-based U-Net. In our case, the dataset used for finetuning is relatively small, making it insufficient to generalize transformer models, which primarily perform well in data-intensive environments due to their low inductive bias~\cite{inductivebias2020dosovitskiy2020image}. Especially in the case of wrinkles, the relationship between adjacent pixels (skin) plays a crucial role in their assessment. Therefore, the CNN-based standard U-Net, which excels at capturing local information, tends to outperform the Swin UNETR, which includes transformer blocks specialized in capturing global context through multi-head attention mechanisms. Nevertheless, our experimental results show that the performance of Swin UNETR progressively improves through our method, suggesting that with more data and longer pretraining, there is significant potential for performance enhancement. Note that accuracy is very high in all experiments since wrinkles occupy a very small proportion of the face and most of the predictions are background pixels.

\begin{figure}[tb]
  \centering
  \includegraphics[width=0.9\textwidth]{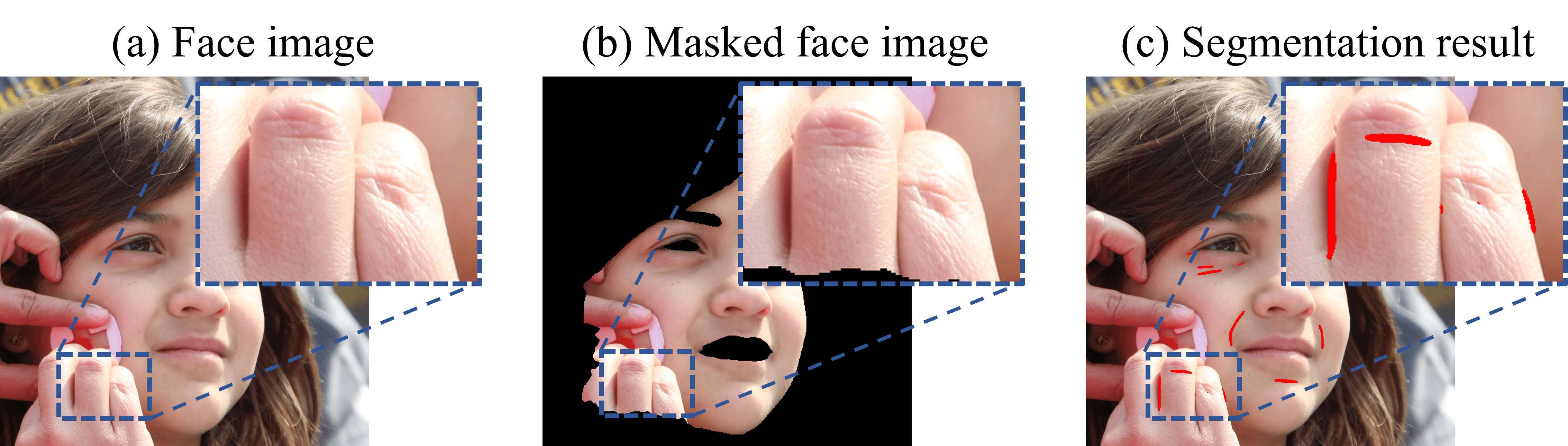}
  \caption{Example of a false wrinkle detection. (a) Face image. (b) Masked face image used as the model input during the finetuning stage. (c) Visualization of the model's predicted segmentation after the finetuning stage.}
  \label{fig:fig10}
\end{figure}

However, our approach has limitations. As shown in Fig. \ref{fig:fig10}, objects similar to wrinkles, such as hair or fingers covering the face, are mistakenly recognized as wrinkles in the images. This results in false positives during the wrinkle segmentation process. To address this issue, upcoming studies will focus on developing techniques that can accurately segment facial regions and precisely distinguish between wrinkle and non-wrinkle areas to reduce false positives. Also, there may be benefits to including the type of wrinkle (e.g., static vs. dynamic wrinkle) to each wrinkle in the facial image because treatment strategies often differ by the type in clinics~\cite{botoxsmall2014botulinum,fillergoldman2011hyaluronic,resurfacinggao2022clinical}. Despite majority voting, the subjectivity in wrinkle annotation remains a challenge. Moving forward, we plan to collaborate with dermatologists for wrinkle annotation and explore techniques such as soft labeling to improve the reliability and trustworthiness of ground truth wrinkles.

\section{Conclusion}
We propose a two-stage learning strategy for facial wrinkle segmentation that leverages transfer learning from facial texture feature extraction. Specifically, the model is pretrained using automatically generated weak wrinkle labels (masked texture maps) to learn general facial features such as contours and skin texture. The model is then finetuned with a smaller set of manually labeled wrinkle data to enhance segmentation performance. This method demonstrates both qualitatively and quantitatively superior results, achieving state-of-the-art performance. Consequently, it significantly reduces the time and cost of manual wrinkle labeling, offering potential benefits in cosmetic dermatology. Additionally, the pretraining method's architecture-independent nature suggests its broad applicability to various segmentation models, making it valuable not only in facial wrinkle segmentation but also in other areas requiring the segmentation of thin objects where manual labeling is costly. To support ongoing research and reproducibility, we have made the FFHQ-Wrinkle dataset—the first publicly available dataset of its kind—accessible to the research community. This dataset comprises 1,000 manually labeled wrinkle images and 50,000 weakly labeled images. By sharing this dataset, we aim to facilitate the development of more advanced wrinkle detection models and promote further advancements in this field.

\subsubsection{Acknowledgements} 
The authors appreciate Dr. Ik Jun Moon, a dermatologist at Asan Medical Center, for sharing invaluable insights and feedback from a dermatological perspective. This work was supported by the National Research Foundation of Korea (NRF) grants funded by the Ministry of Science and ICT (MSIT) (RS-2024-00455720 \& RS-2024-00338048), the National Institute of Health(NIH) research project (2024ER040700), the National Supercomputing Center with supercomputing resources including technical support (KSC-2024-CRE-0021), Hankuk University of Foreign Studies Research Fund of 2024, the artificial intelligence semiconductor support program to nurture the best talents (IITP(2024)-RS-2023-00253914) grant funded by the Korea government, and the Culture, Sports and Tourism R\&D Program through the Korea Creative Content Agency grant funded by the Ministry of Culture, Sports and Tourism in 2024(RS-2024-00332210).

%
%
%
\bibliographystyle{splncs04}
\bibliography{main}
\end{document}